\documentclass{article}

\usepackage{PRIMEarxiv}
\usepackage[utf8]{inputenc}
\usepackage[T1]{fontenc}
\usepackage{microtype}
\usepackage{xcolor}
\usepackage{hyperref}
\usepackage[round]{natbib}
\usepackage{url}
\usepackage{booktabs}
\usepackage{graphicx}
\usepackage{tabularx}
\usepackage{array}
\graphicspath{{media/}}

\newcolumntype{Y}{>{\raggedright\arraybackslash}X}
\definecolor{darkblue}{rgb}{0,0,0.5}
\hypersetup{colorlinks=true,citecolor=darkblue,linkcolor=darkblue,urlcolor=darkblue}

\title{Artifact-centered Claim-aware Observability for Autonomous Scientific Agents}

\author{
  Xiangyu Yin$^{*}$, Ming Du, Michael H. Prince, and Mathew J. Cherukara$^{\dagger}$ \\
  Advanced Photon Source, Argonne National Laboratory, Lemont IL, USA \\
  $^{*}$\texttt{xyin@anl.gov}; $^{\dagger}$\texttt{mcherukara@anl.gov} \\[0.75em]
  \normalfont Accepted by LM4Sci Workshop at the Conference on Language Modeling (COLM) 2026.
}

\begin{document}
\pagenumbering{gobble}
\thispagestyle{empty}
\textbf{GOVERNMENT LICENSE}

The submitted manuscript has been created by UChicago Argonne, LLC, Operator of Argonne National Laboratory (“Argonne”). Argonne, a U.S. Department of Energy Office of Science laboratory, is operated under Contract No. DE-AC02-06CH11357. The U.S. Government retains for itself, and others acting on its behalf, a paid-up nonexclusive, irrevocable worldwide license in said article to reproduce, prepare derivative works, distribute copies to the public, and perform publicly and display publicly, by or on behalf of the Government. The Department of Energy will provide public access to these results of federally sponsored research in accordance with the DOE Public Access Plan. \href{http://energy.gov/downloads/doe-public-access-plan}{http://energy.gov/downloads/doe-public-access-plan}
\clearpage
\pagenumbering{arabic}

\maketitle

\begin{abstract}
Autonomous scientific agents now increasingly propose ideas, write code, run experiments, analyze results, and even draft papers. Observe and audit those agents are necessary but logging every model call is not enough, scientists also need to inspect the artifacts and claims that the systems produced and their relations. This is driven by the fact that failures in scientific agent systems are often distributed across several objects. A manuscript claim may cite the wrong evidence, a search process may select a degenerate candidate, a laboratory novelty claim may depend on an unstated rule, or a multi-agent plan may change without a visible trigger. Existing tracing, experiment tracking, and archival provenance tools are valuable, but their native objects do not make these scientific audit relations first-class. We argue that autonomous scientific systems should emit portable, claim-aware artifact lineage as a minimum audit layer. We propose a compact observability profile organized around individuals, operators, fitness records, lineage, archives, runs, streams, and steering commands. In this profile, scientific claims are ordinary individuals with explicit evidence bindings and verification records. The profile is intended as a semantic layer that complements current telemetry and provenance standards. Execution details can remain in OpenTelemetry. Final packages can export to PROV-O or RO-Crate standards.
\end{abstract}

\section{The audit unit is the artifact}

Autonomous scientific agents are moving from assistants that answer questions toward systems that help execute research workflows. Recent systems show they can automate paper writing pipelines, agentic tree search over research ideas, code generation for experiments, literature synthesis, laboratory planning, and closed-loop chemistry or materials workflows~\citep{lu2026endtoend,yamada2025aiscientistv2,schmidgall2025agentlab,romeraparedes2024funsearch,alphaevolve2025,boiko2023coscientist,bran2024chemcrow,swanson2025virtuallab}. This shift changes what it means to observe a scientific computation system. A log of prompts, completions, tool calls, and timestamps is necessary, but it is not sufficient. A scientist also needs to know which artifact the agent produced, which earlier artifacts led to it, what evidence supports each claim, which evaluator accepted or rejected it, and which branch or human intervention changed the trajectory.

We argue here that autonomous scientific agentic systems should make artifact lineage and claim-evidence bindings first-class observability records. Span trees and run logs should not be the only portable audit trail. The minimum trace for a scientific agent should include candidate artifacts, the operators that derive them, evaluator outputs attached to those artifacts, archive or selection decisions, human steering events, and claims represented as inspectable artifacts with evidence bindings.

This distinction matters because the most damaging failures of scientific agents are often not localized to a single model/tool call. Audits of manuscripts produced by automated research have reported coding failures, hallucinated or internally inconsistent numerical results, and malformed or unsupported claims~\citep{beel2025evaluating,xu2026ghostcite}. In autonomous materials synthesis system, novelty and yield claims have required later correction or dispute after the system completed its loop~\citep{szymanski2023autonomous,leeman2024challenges,szymanski2026correction}. Such cases require an audit trail that connects claims with evidence, evaluators, and lineage. A span tree can tell us which calls occurred. It does not by itself identify the scientific claim, bind the claim to a measurement, or show that a later draft reused an invalidated result.

\begin{figure}[t]
\centering
\includegraphics[width=\linewidth]{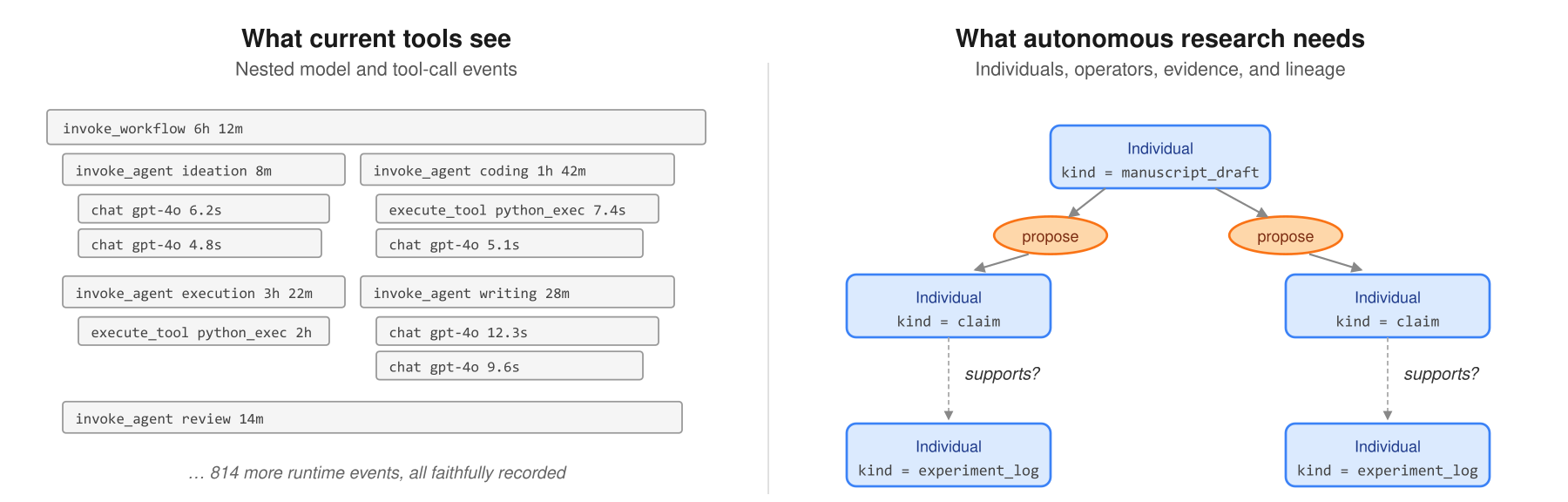}
\caption{The observability shift. Runtime spans record nested calls and timing. Claim-aware artifact observability records candidate artifacts, the operators that derive them, evaluator outputs, archive decisions, and evidence links for claims.}
\label{fig:hero}
\end{figure}

Figure~\ref{fig:hero} summarizes the shift. The left side resembles a conventional trace: an ordered tree of calls. The right side shows the scientific units of inspection: artifacts, lineage, evaluations, and evidence. Our proposal is not to replace tracing, experiment tracking, or archival provenance, but to add the missing semantic layer that lets autonomous research systems answer scientific audit questions during a run and compile richer archival packages afterward.

\section{Why current infrastructure can miss scientific audit objects}

Current infrastructure provides several partial views. Tracing systems for LLM applications record spans: prompts, completions, tool invocations, latency, token counts, and nested call structure~\citep{langsmith,langfuse,phoenix,wandb}. Experiment trackers record runs, parameters, metrics, artifacts, and sometimes model lineage~\citep{mlflow,wandb}. Provenance and research object standards such as W3C PROV-O, PROV-DM, and RO-Crate record entities, activities, agents, derivations, and bundled research artifacts for later sharing and reproducibility~\citep{provo,provdm,rocrate}. Version control systems for code and data, such as git, git-LFS, and DVC, track lineage as content-addressed commit graphs with branch and merge structure~\citep{dvc}. These tools are useful, mature, and should be reused.

The gap is a mismatch of primary units. A span-centric system asks, ``which call happened next?'' An experiment tracker asks, ``which run produced this metric?'' An archival package asks, ``how was this final object derived?'' A version control system asks, ``which lines changed between these two file trees?'' A scientist auditing an autonomous system asks different questions: which candidate contains this claim, what evidence did it read, which evaluator accepted it, and what branch selected it? Those relations can be encoded in tags, metadata, file paths, or commit messages, but the conventions are local and fragile. Two teams can both log every model call and still be unable to exchange traces that answer the same query about claims.

Three audit scenarios illustrate the problem. First, an agent that writes papers may report a numerical improvement that contradicts the experiment log. The audit object is not the prompt that wrote the sentence. It is the claim, the evidence artifact it cites, the extracted value, and the verification result. Second, an evolutionary or tree-search system may appear to improve while the population collapses to near-duplicates that exploit a benchmark quirk. The audit object is the lineage and fitness of the candidate population, not any single model completion. Third, a closed-loop lab may emit delayed physical measurements, sample identifiers, instrument files, and operator approvals. The audit object spans software decisions and physical evidence, and may not align with synchronous call boundaries.

Metadata conventions are not enough for scientific review. A parent program ID can be stored as a span attribute, a token in a commit message, a database row, or a file name. A claim's evidence can be stored as a citation string, a notebook path, a retrieval snippet, or a table coordinate. All are workable inside one lab. None gives a reviewer, benchmark, or downstream archive an easy and portable way to ask common audit questions: show every unsupported claim, trace a result to its evidence, find the branch where a human intervention changed the plan, or list archive members that were later reused in the manuscript.

\section{Audit questions reveal observability requirements}

Claim-aware artifact observability starts from the questions that scientists, reviewers, and safety monitors need to ask. These questions mix distinct relation types. Some ask for provenance, such as which parent artifact produced a child. Some ask for epistemic support, such as which measurement or literature item supports a claim. Some ask for control history, such as which human instruction or automated policy changed the search. An observability contract should keep these relations distinct rather than compressing them into a single generic dependency edge.

\begin{table}[t]
\centering
\small
\begin{tabularx}{\linewidth}{p{0.28\linewidth}YY}
\toprule
Audit question & Required relation & Example failure exposed \\
\midrule
Where did this result come from? & Parent individuals, derivation operator, emitter stream & Final candidate descends from a failed or out-of-policy branch \\
What supports this claim? & Claim individual, evidence references, evidence bindings & Manuscript cites a table whose value contradicts the sentence \\
Who or what accepted it? & Fitness record, evaluator identity, threshold or rubric & One unstable evaluator dominates a multi-evaluator pipeline \\
How was the search steered? & Steering command, applied operator, affected archive or plan & Human correction is overwritten by a later autonomous step \\
What was discarded? & Archive-update and rejection events & Negative results or failed replications disappear from the record \\
\bottomrule
\end{tabularx}
\caption{Design requirements derived from scientific audit questions. The profile separates derivation, evidence, evaluation, and steering because each supports a different class of review.}
\label{tab:audit}
\end{table}

Table~\ref{tab:audit} also explains why simple file provenance is not enough. A generated sentence can be derived from a draft without being supported by the data it mentions. A code candidate can descend from a strong parent while its reported score comes from an unstable evaluator. A plan revision can be causally downstream of a human instruction while still failing to preserve the user's intended constraint. The observability schema must therefore represent derivation, evidence, fitness, and steering as separately queryable facts.

This matters most when reviewers assess scientific agent outputs. A reviewer may not need to replay the whole run. They may need to inspect a few high-risk claims: numerical comparisons, novelty assertions, safety-relevant experimental decisions, or literature claims that could be hallucinated. A trace format that exposes these objects makes review targeted rather than forensic. Instead of reading thousands of spans, a reviewer can ask for all final claims with missing evidence, all accepted candidates with no independent evaluator, or all archive decisions made after an intervention.

\section{Claim-aware observable event trace profile}

A trace is an append-only event log. Derived views (current archive membership, best lineage path, evaluator dashboards, claim graphs, and worker utilization) are computed from the event trace sequence. This design avoids treating the latest dashboard state as the scientific record. If a system removes a candidate from an archive, revises a claim, or changes a plan, the event that made the change remains inspectable.

The proposed profile has five scientific abstractions and three structural records. An \textbf{individual} is any candidate artifact under inspection: a program, manuscript draft, table, protocol, plan, sample, molecule, figure, claim, or agent version. An \textbf{operator} is an event that derives one or more child individuals from zero or more parents, or acts on an existing record. A \textbf{fitness} record is an evaluator output attached to a specific individual: a scalar score, pass/fail result, structured review, proof-check status, wet-lab measurement, novelty classification, or claim-verification judgment. \textbf{Lineage} is the parent-child graph induced by operators. An \textbf{archive} is an optional curated set, such as an elite set, a Pareto front, a set of accepted claims, a list of retained samples, or a MAP-Elites grid. Three structural records complete the profile. A \textbf{run} frames one autonomous process. A \textbf{stream} identifies a worker, agent role, instrument, tool subsystem, or human interface. A \textbf{steering command} records a human or external intervention and links, when applied, to the operator that changed the run.

The profile is deliberately small (Appendix~\ref{app:contract} gives the event contract and structural invariants). It does not prescribe how to store a microscope image, diffraction pattern, molecule graph, proof object, code repository, or manuscript. Instead, it requires every domain payload to be addressable by reference and hash, and every scientifically relevant transformation to be represented as an operator over individuals. The result is a portable query surface: parentage, evidence, evaluation, selection, and intervention have stable meanings even when payloads are domain-specific.

\subsection{Claims as individuals}

A central requirement for scientific agent observability is claim inspection. A generated paper, report, or lab note carries a particular risk: unsupported claims can look polished and pass through downstream review even when they are wrong. We therefore treat a scientific claim as an ordinary individual rather than a special annotation embedded inside a manuscript. A claim individual has text or structured content, evidence references, optional evidence bindings, and fitness records produced by verifiers.

Evidence references list the artifacts the claim points to: measurements, tables, code outputs, literature items, human reviewer notes, instrument files, or earlier claims. Evidence bindings optionally record how the evidence was used: which field was read, which value was extracted, what comparison was made, and why that evidence supports or weakens the claim. A verifier then emits a fitness record such as \texttt{supported}, \texttt{unsupported}, \texttt{contradicted}, or \texttt{needs\_human\_review}. Later manuscript drafts inherit, revise, or reject claim individuals through ordinary operators.

This design makes failures in paper writing queryable. A reviewer or operator can ask for all claims in the final draft whose evidence does not exist, whose extracted value differs from the cited metric, or whose verifier failed. It also avoids a separate system for claim provenance. Claims can be selected into archives, superseded by new evidence, or traced back to earlier drafts just like other artifacts. Because claim verification can be partial, the profile does not assume perfect verifiers. It only makes the verification target and result explicit.

A claim individual is not a guarantee of truth. It is a stable handle for review. Verifiers may be automatic scripts, retrieval-augmented checkers, domain models, theorem provers, human reviewers, or later laboratory measurements. The important invariant is that the claim, its evidence, and the verification result remain separately identifiable. This lets a later system invalidate a claim without rewriting history. The old claim remains in the log, a new fitness record captures the contradiction, and a revision operator can produce a corrected claim or remove it from the accepted archive.

\subsection{Interoperability rather than replacement}

Claim-aware artifact observability should be a bridge, not a replacement. Execution details can remain in OpenTelemetry GenAI semantic conventions~\citep{otelgenai}. An operator can reference one or more spans as low-level evidence for how the event was executed. Final derivation graphs can export to PROV-O by mapping individuals to entities, operators to activities, streams or humans to agents, and parent-child edges to usage/generation relations~\citep{provo,provdm}. Research packages can serialize payloads and metadata through RO-Crate~\citep{rocrate}. Workflow datasets can export coarse-grained input/output relations through OpenLineage~\citep{openlineage}. Nothing in the profile fixes the underlying store. A version control system can serve as its substrate: git already supplies content-addressed identifiers and a multi-parent derivation graph, and branch, tag, and commit conventions can encode individuals, operators, fitness records, and archive decisions. It adds a standard schema rather than replacement, so those conventions become portable across labs rather than local to one repository.

The profile is also complementary to the emerging ideas such as agent-native research (ARA) artifacts. \citet{liu2026ara} argue that traditional papers hide failed branches and omit engineering detail, and they propose agent-native research artifacts as packages containing logic, code, exploration, and evidence. Our proposal records the run from which such a package can be compiled. An agent-native artifact is an end-state package. Claim-aware artifact observability is the live log that preserves branch structure, evaluator history, claim evidence, and steering events before the final package is assembled.

\section{Proposed requirements for the community}

We understand fully adopting the proposed observability profile can be infeasible in short terms, so we propose five minimum requirements for autonomous scientific agents for research venues, benchmarks, and tool builders such that they can still benefit from the ideas of this proposal.

\textbf{Emit artifact IDs, not only span IDs.} Every audit-relevant research object created by an agent should have a stable individual ID, kind, payload reference or hash, creator operator, timestamp, and status. The right granularity depends on the domain, but final claims, tables, figures, experiment logs, code versions, plans, and selected candidates should never exist only inside prompt text.

\textbf{Emit operators as lineage events.} Any refinement, mutation, synthesis, extraction, review, verification, selection, or plan update should record its parent individuals, child individuals, stream, tool or agent, and parameters sufficient to interpret the edge. This makes the lineage graph replayable and queryable.

\textbf{Attach evaluator outputs to artifacts.} Metrics should not float at the run level when they judge a specific artifact. A benchmark score belongs to a code candidate, a novelty decision belongs to a sample or claim, and a reviewer score belongs to a draft or claim set.

\textbf{Represent claims and evidence explicitly.} Scientific statements that appear in generated reports should be extracted or emitted as claim individuals. Each claim should link to evidence artifacts and, when possible, to the exact value or field used. Verification is then a normal evaluation step.

\textbf{Keep human steering and archive decisions in the log.} Pauses, approvals, forced forks, rejected candidates, archive additions, plan revisions, and safety stops should be visible events, not dashboard state that disappears when a run completes.

These requirements leave implementation choices open. They do not force one database or one UI. They ask systems to expose the units that scientific review already depends on. Appendix~\ref{app:queries} lists representative audit queries, and Appendix~\ref{app:tracelets} gives compact tracelets for common failure modes. A reviewer should be able to load a trace and ask which claims are unsupported, which artifact introduced a result, which evaluation accepted it, which human command changed the path, and which archive members were later reused in the manuscript.

\section{A minimal worked example}

Consider a common failure: an agent writes that a new method improves accuracy by 4.2 percentage points, but the experiment log shows a smaller gain or no gain. In a span-only trace, the claim may be buried in a completion and the evidence may be a file path in a tool output. In the proposed profile, the same event sequence can be represented as a small graph of records.

\begin{table}[t]
\centering
\caption{A compact trace for an unsupported numerical claim. The payloads can live outside the trace. The trace preserves the audit relations.}
\label{tab:worked}
\small
\begin{tabularx}{\textwidth}{@{}p{1.0cm}p{2.35cm}Y@{}}
\toprule
Step & Record & Audit-relevant fields \\
\midrule
1 & Individual: experiment log & \texttt{id=i\_exp}, \texttt{kind=experiment\_log}, payload hash, metric table URI. \\
2 & Individual: draft & \texttt{id=i\_draft}, \texttt{kind=manuscript\_draft}, payload hash for generated text. \\
3 & Operator: extract claim & Parent \texttt{i\_draft}, child \texttt{i\_claim}, stream \texttt{writer}, extractor version. \\
4 & Individual: claim & \texttt{kind=claim}, text payload, \texttt{evidence\_refs=[i\_exp]}, binding to metric row and value. \\
5 & Operator: verify & Parent \texttt{i\_claim}, context \texttt{i\_exp}, evaluator \texttt{claim\_numeric\_checker}. \\
6 & Fitness: verification & Target \texttt{i\_claim}, value \texttt{unsupported}, reason \texttt{evidence\_mismatch}, extracted and claimed values. \\
\bottomrule
\end{tabularx}
\end{table}

Table~\ref{tab:worked} simplifies the inspection workflow. A reviewer does not need to read every model call to find the problem. They can query claim individuals whose verification fitness is unsupported or contradicted, inspect the evidence binding, and follow lineage back to the draft and extraction operator. A developer can still open the corresponding spans to debug the prompt or tool. A provenance archive can still export the final graph. The profile connects these views through stable artifact IDs.

The same pattern scales to other domains. A novelty claim can bind to a sample, diffraction pattern, database entry, and decision threshold. A proof claim can bind to a theorem statement and proof-checker output. A claim in protein design can bind to a structure prediction, assay result, and acceptance rule. In each case, the scientific assertion is not treated as free text. It is a research artifact with provenance, evidence, and evaluation.

\section{Trace shapes and example encoding}

\begin{figure}[t]
\centering
\includegraphics[width=\linewidth]{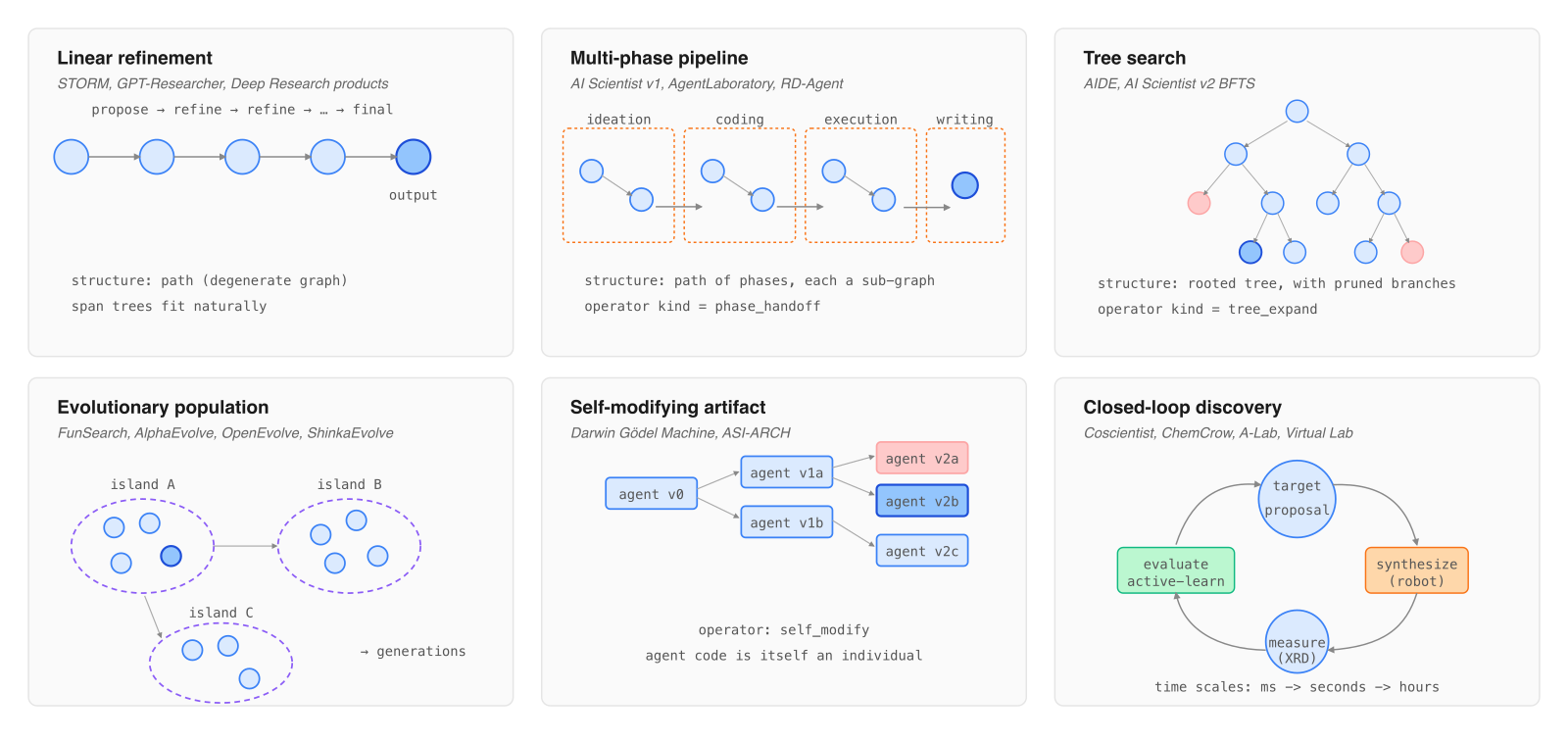}
\caption{Six trace shapes that can share the same artifact-centered vocabulary: linear refinement, multi-phase pipeline, tree search, evolutionary population, self-modifying artifact, and closed-loop discovery.}
\label{fig:shapes}
\end{figure}

The proposed profile can cover more than a simple agent chat loop. Autonomous research traces often appear as linear refinement, multi-phase pipelines, tree search, evolutionary populations, self-modifying artifacts, or closed-loop discovery processes (Figure~\ref{fig:shapes}). The control structure changes, but the audit units remain the same: artifacts, derivations, evaluations, selections, evidence links, and interventions.

\textbf{Linear refinement and multi-phase writing.} A system that writes reports can emit a draft individual at each revision, extraction operators for numerical or literature claims, and verifier fitness records after checking claims against experiment outputs or cited sources. A multi-phase pipeline adds typed handoff artifacts (research questions, code repositories, experiment logs, analysis summaries, drafts, and internal reviews) so a final claim can be traced through the phase that introduced it and the evidence that supports it.

\textbf{Tree search and evolutionary populations.} A tree-search system represents each state, plan, or candidate draft as an individual and each expansion as an operator. An evolutionary design system represents candidate programs, molecules, or hypotheses as individuals. Mutation and crossover are operators. Benchmark, simulation, or assay results are fitness records, and survivor sets or Pareto fronts are archive events. These records expose whether improvement comes from diverse lineages or from collapse onto near-duplicate candidates that exploit one evaluator.

\textbf{Self-modifying and multi-agent systems.} In self-modifying systems, prompts, policies, evaluator definitions, and agent code are themselves individuals. A self-modification is an operator that derives a new agent version. Later operators list that version in context, so the event log remains acyclic even when the system changes the process that creates future artifacts. In a hub-and-spoke research team, streams distinguish a PI, coder, runner, critic, instrument, or human supervisor. Message and plan revision operators then show which specialist artifact or human intervention changed the shared plan.

\textbf{Closed-loop laboratories.} A laboratory workflow can represent protocols, samples, instrument outputs, measurements, and novelty or yield claims as individuals. Delayed measurements become evidence or fitness records even if they arrive long after the planning span has ended. If a later manuscript claims that a synthesis improved yield or discovered a novel material, the trace can connect the claim to the relevant sample identifiers, instrument files, extraction operator, and acceptance criterion.

These examples do not require separate schema. They require the same small set of records emitted beside existing logs and domain-specific payloads referenced by hash or URI. The live system can keep its own databases and orchestration runtime. The artifact-centered log supplies the exchange layer for common review queries: the lineage of the best candidate, unsupported claims, evaluator disagreement, archive churn, the cause of a plan revision, and the effect of a human intervention.

\section{Discussion \& Limitations}

The profile can be tested against existing review practices. It also raises practical questions about deployment overhead, integrity, and scope.

\textbf{Audit coverage.} Given traces from pipelines that generate papers, evolutionary code search, multi-agent planning, and closed-loop experiments, independent auditors should be able to recover the evidence for each reported claim, the operator that introduced it, and the evaluation that accepted or rejected it. The baseline should be the best available existing records, not an intentionally weak comparison.

\textbf{Reviewer utility.} Generated manuscripts can be reviewed with and without claim-aware artifact traces. Useful outcome measures include time to identify unsupported claims, agreement among reviewers about failure causes, and the number of reconstruction steps performed outside the trace. The profile is useful only if it reduces review burden rather than adding another artifact reviewers ignore.

\textbf{Trace benchmarks.} A useful benchmark would pair research artifacts generated by agents with gold audit labels: claim records, evidence objects, verifier outcomes, lineage edges, archive decisions, and steering events. A minimal package for reviewers could include a schema file, a redacted or synthetic run log, a manifest mapping payload hashes to public artifacts, validator output, and query examples matching Table~\ref{tab:audit}. This is weaker than full reproducibility, but stronger than a prose assurance that a system was logged.

\textbf{Overhead and portability.} A practical profile must tolerate high-frequency tool calls without forcing every token or scratch file into the lineage graph. Tiered logging offers one path: cheap IDs and hashes for most artifacts, richer evidence bindings for selected claims, and domain-specific expansion when an audit requires it. The same run should also export to OpenTelemetry, PROV-O, RO-Crate, and agent-native archives without losing claim, lineage, fitness, archive, or steering relations (Appendix~\ref{app:package} outlines one such package).

\textbf{Integrity and redaction.} High-stakes deployments need tamper evidence as well as privacy controls. Append-only event streams, event signatures, hash chains, payload redaction records, and access policies can help distinguish a missing payload from a missing event. These mechanisms do not make a bad evaluator correct, but they make silent rewriting of the scientific record harder and clarify which parts of the trace were emitted online versus reconstructed after the run.

\textbf{Other Limitations.} Artifact-centered observability does not prevent hallucination, fabrication, unsafe laboratory action, or benchmark overfitting. It makes the objects needed to detect and investigate such failures explicit. Verification quality still depends on evaluators, domain rules, and human judgment. A trace can faithfully record a bad novelty criterion or a biased benchmark. The point is that the criterion and benchmark become visible objects. The hardest open question is granularity: each domain needs conventions for what counts as a material individual, especially when claim extraction is partly automated and partly post hoc.

\section{Conclusion}

Autonomous science needs observability at the level of scientific artifacts. The relevant audit object is often not a prompt, span, or run, but a candidate artifact and the relations among claims, evidence, and evaluation around it. We argue that artifact-centered, claim-aware traces should become a minimum layer for scientific agent systems. The profile proposed here is small by design: individuals, operators, fitness records, lineage, archives, runs, streams, and steering commands. It complements existing telemetry and provenance standards while giving reviewers the graph they need to inspect generated research. 

\section*{Acknowledgments}
This research used resources of the Advanced Photon Source, a U.S.~Department of Energy (DOE) Office of Science user facility at Argonne National Laboratory, and is based on research supported by the U.S. DOE Office of Science-Basic Energy Sciences, under Contract No.~DE-AC02-06CH11357.

\clearpage
\bibliography{references}
\bibliographystyle{unsrtnat}

\clearpage
\appendix
\makeatletter
\@addtoreset{figure}{section}
\@addtoreset{table}{section}
\makeatother
\renewcommand{\thefigure}{\Alph{section}.\arabic{figure}}
\renewcommand{\thetable}{\Alph{section}.\arabic{table}}
\renewcommand{\theHfigure}{appfig.\arabic{section}.\arabic{figure}}
\renewcommand{\theHtable}{apptab.\arabic{section}.\arabic{table}}

\section{Event contract and structural invariants}\label{app:contract}

\begin{figure}[h]
\centering
\includegraphics[width=0.90\linewidth]{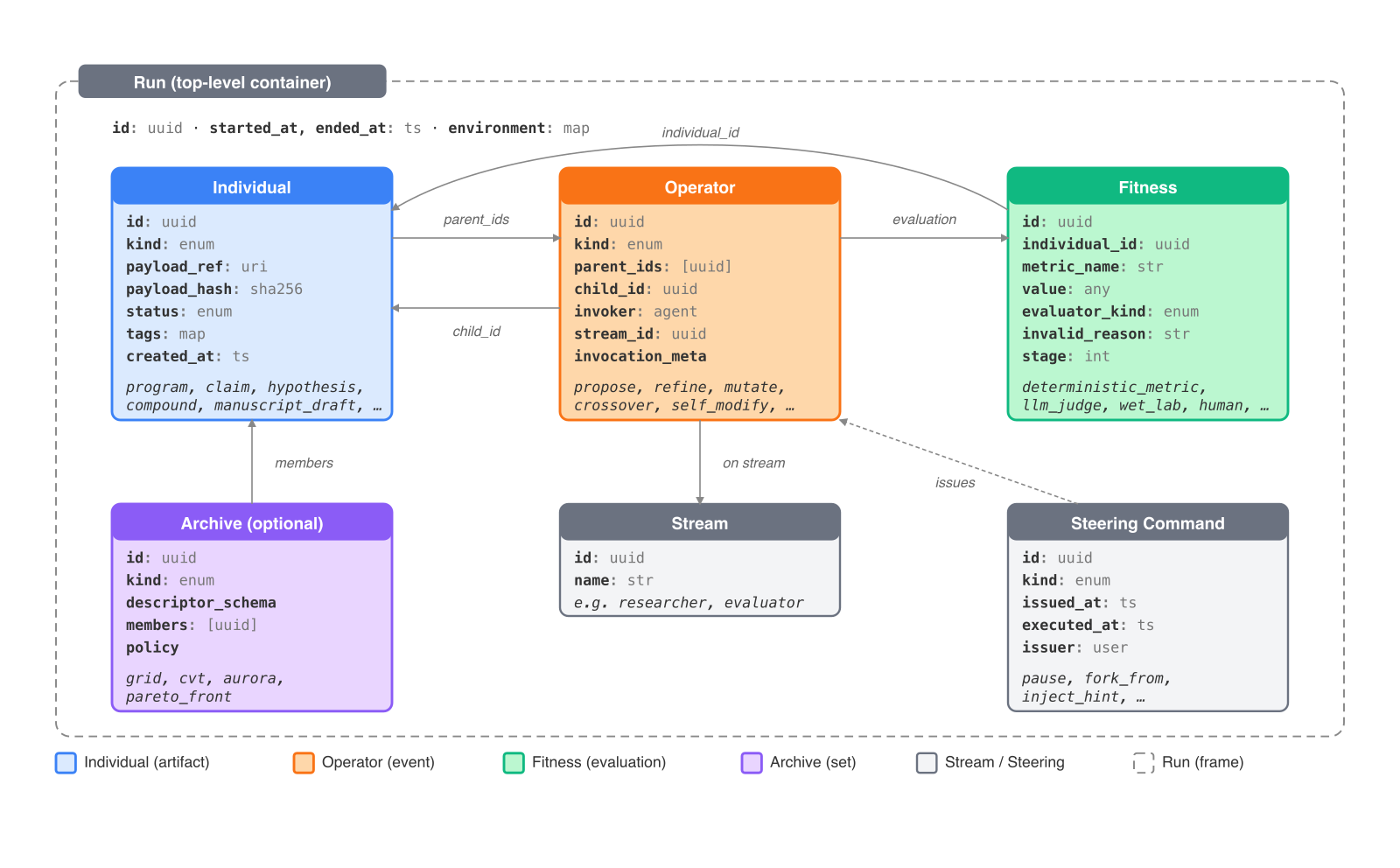}
\caption{Core records and their reference structure. Payloads can remain in notebooks, databases, repositories, instrument stores, or archives. The trace records the identities and relations needed for audit.}
\label{fig:schema_app_clean}
\end{figure}

The trace profile uses a compact event vocabulary. Figure~\ref{fig:schema_app_clean} shows the core records and their reference structure. Table~\ref{tab:event_contract} records the fields that must survive export for the trace to remain inspectable. Domain payloads can live outside the event stream. The event stream only needs stable identifiers, typed relations, hashes or URIs, and enough version information to reconstruct the audit graph.

\begin{table}[h]
\centering
\scriptsize
\begin{tabularx}{\linewidth}{p{0.20\linewidth}Y Y}
\toprule
Record & Minimum fields & Integrity rule \\
\midrule
Individual & \texttt{id}, \texttt{run\_id}, \texttt{kind}, \texttt{created\_at}, payload reference or hash & The identifier denotes one inspectable artifact version: draft, code, dataset, sample, measurement, plan, policy, evaluator, or claim. \\
Operator & \texttt{id}, \texttt{op\_type}, \texttt{parent\_ids}, \texttt{child\_ids}, emitter stream, timestamp & Children are derived from earlier parents or explicitly imported external objects. Context references do not imply derivation. \\
Fitness record & evaluated individual, evaluator identity and version, value or verdict, timestamp & The evaluator version is part of the result. A later evaluator may disagree without overwriting the earlier record. \\
Evidence binding & claim identifier, evidence identifier, locator, extracted value or rule, binding role & A claim can cite many evidence objects, and each citation can fail independently. \\
Archive event & archive identifier, member identifier, add/remove action, rule or policy, timestamp & Membership history is event-sourced. The current archive is a derived view. \\
Run and stream & run identifier, stream identifier, invoker identity, parent run when applicable & Parallel agents, tools, humans, and instruments can emit into separate streams without requiring separate schemas. \\
Steering command & command identifier, issuer, target stream or operator, instruction type, effect reference & Human interventions and policy updates are auditable artifacts, not comments hidden in logs. \\
\bottomrule
\end{tabularx}
\caption{A minimal event contract for portable artifact-centered traces.}
\label{tab:event_contract}
\end{table}

Five invariants are especially important. First, every reference either resolves to an earlier event object or is marked as an external import. Second, derivation edges are acyclic by artifact version, even when the system modifies its own code, prompt, or evaluator. Third, every final scientific claim has an evidence binding, a failed verification record, or an explicit unsupported status. Fourth, evaluator outputs name the evaluator version and the individual they judged. Fifth, redaction removes payload access but not the fact that an event existed. Otherwise a reviewer cannot distinguish private evidence from missing evidence.

\section{Inspection queries}\label{app:queries}

The trace is useful only when common audit questions can be answered mechanically. Table~\ref{tab:queries_app} gives representative queries and the records they require. These queries are deliberately stated at the schema level rather than in a particular database language, because an implementation may store events in JSONL, SQL tables, graph databases, version control history, or an experiment tracker with exported facets.

\begin{table}[h]
\centering
\scriptsize
\begin{tabularx}{\linewidth}{p{0.27\linewidth}Y Y}
\toprule
Query & Records read & Expected answer \\
\midrule
List unsupported final claims & Final report archive, claim individuals, evidence bindings, verification fitness & Claims with no supporting evidence, failed verification, stale evidence, or explicit unsupported status. \\
Trace a number in the manuscript & Claim individual, locator, evidence binding, source table or run artifact, verifier record & The evidence object, extracted value, claimed value, and pass/fail status. \\
Recover the best candidate's ancestry & Archive event, selected individual, lineage operators, parent individuals, fitness records & The branch or population history that produced the selected artifact. \\
Detect archive collapse & Archive events, lineage graph, payload hashes, diversity metrics when present & Whether high-scoring candidates share recent ancestors, duplicate payloads, or one evaluator loophole. \\
Find the cause of a plan revision & Plan individuals, revision operators, context references, message operators, steering commands & The result, reviewer artifact, tool failure, or human intervention that changed the plan. \\
Audit self-modification & Agent-policy individuals, self-modification operators, later operator context & Which agent version produced each result and whether evaluator or policy changes preceded acceptance. \\
Audit closed-loop claims & Sample, protocol, instrument, measurement, extraction, claim, and verifier records & Which physical evidence and decision rule support each novelty, yield, or safety claim. \\
Check redaction boundaries & Payload manifest, redaction records, event hashes, access labels & Which payloads are hidden, why they are hidden, and whether their event identities remain intact. \\
\bottomrule
\end{tabularx}
\caption{Schema-level audit queries enabled by the event contract.}
\label{tab:queries_app}
\end{table}

These queries are a useful acceptance test for an implementation. A system that logs all prompts and tool calls but cannot answer them has execution telemetry, not artifact-centered scientific observability.

\section{Compact tracelets for common failures}\label{app:tracelets}

The following tracelets are intentionally partial (Table~\ref{tab:tracelets}). They show the few records that make an error inspectable, not the full runtime history.

\begin{table}[h]
\centering
\scriptsize
\begin{tabularx}{\linewidth}{p{0.20\linewidth}Y Y}
\toprule
Case & Minimal event sequence & What becomes inspectable \\
\midrule
Hallucinated number & \texttt{draft\_4} $\rightarrow$ \texttt{extract\_claim} $\rightarrow$ \texttt{claim\_23}. \texttt{claim\_23} binds to \texttt{run7\_metrics}. \texttt{metric\_checker\_v2} emits \texttt{fail} with observed value $84.7$ rather than claimed value $87.4$. & The sentence is no longer buried in prose. The trace identifies the claim, the evidence table, the extracted value, the checker version, and the repair or propagation path. \\
Evolutionary collapse & \texttt{program\_87} and \texttt{program\_91} produce children through mutation and crossover. Benchmark fitness rises. Archive events admit near-duplicate payload hashes under one selection rule. & Improvement can be separated from diversity. A reviewer can see whether the archive exploited a benchmark corner rather than discovering robust alternatives. \\
Plan revision & A runner artifact reports a failed experiment. A critic stream emits a review. A human steering command changes the priority. \texttt{revise\_plan} derives \texttt{plan\_v2} from \texttt{plan\_v1} with both records as context. & The plan change has a visible cause. Cross-agent handoff and human steering are part of lineage rather than chat transcript residue. \\
Closed-loop novelty & \texttt{sample\_s17}, \texttt{protocol\_p4}, and \texttt{xrd\_file\_s17} feed an extraction operator. A novelty rule emits a verdict. A manuscript claim binds to the rule, reference set, and instrument file. & A later correction can identify whether the disputed claim came from the measurement, the extraction, the database comparison, or the novelty rule. \\
Self-modification & \texttt{agent\_v2} edits its prompt and evaluator definition, producing \texttt{agent\_v3}. Later candidate-generation operators list \texttt{agent\_v3} as context. & The trace remains acyclic by version while showing that the process generating later artifacts changed. \\
\bottomrule
\end{tabularx}
\caption{Small trace fragments for diagnosing common audit failures.}
\label{tab:tracelets}
\end{table}

\section{Packaging, export, and redaction}\label{app:package}

A reviewable trace package does not require access to a private production system. The portable object can be small: an event stream, a payload manifest, validator output, and a few inspection queries (Table~\ref{tab:package}). The package should separate event identity from payload access so that private data can be redacted without erasing the scientific dependency graph.

\begin{table}[h]
\centering
\scriptsize
\begin{tabularx}{\linewidth}{p{0.24\linewidth}Y Y}
\toprule
Package component & Contents & Failure mode avoided \\
\midrule
Event stream & JSONL, database export, or graph export containing the contract in Table~\ref{tab:event_contract} & Reviewers receive files but cannot reconstruct relations among claims, evidence, and evaluations. \\
Payload manifest & Hashes or URIs for drafts, code, tables, datasets, notebooks, instrument files, figures, and external references & A trace points to objects whose identity changed after the run. \\
Validator report & Reference checks, acyclicity checks, required-field checks, final-claim coverage, redaction accounting & Missing or malformed relations are discovered only during manual review. \\
Inspection queries & Saved queries for unsupported claims, lineage, archive history, steering effects, and redaction boundaries & Each reviewer reconstructs a different audit procedure. \\
Telemetry export & Span identifiers, model metadata, tool calls, latency, token counts, errors, and links to produced individuals & Execution debugging is separated from scientific artifact audit. \\
Archival export & PROV-O, RO-Crate, OpenLineage, or domain archive bundle when applicable & Final research objects lose the running history that explains how they were selected. \\
\bottomrule
\end{tabularx}
\caption{A compact package for review, benchmarking, or later archival export.}
\label{tab:package}
\end{table}

Export to existing standards should preserve artifact identifiers. Operators can link to OpenTelemetry spans or span groups. In PROV, individuals map to entities, operators to activities, and streams or invokers to agents. Payloads, manifests, validator reports, and query examples can be packaged as RO-Crate objects, and operators that produce data can expose OpenLineage facets when the payload is tabular. The lossy boundary is claim semantics: evidence bindings, verifier versions, archive membership, and steering effects should be preserved as explicit fields or companion files rather than compressed into free text metadata.

\section*{Ethics Statement}
This work advocates audit infrastructure for autonomous scientific agents. It does not introduce a new autonomous agent or deploy a system in the world. The intended ethical benefit is improved accountability for generated scientific claims, human steering decisions, and evidence use. The same trace data could expose sensitive laboratory, personnel, or proprietary information, so deployments should add access control, redaction, and signing policies appropriate to the domain.

\section*{Use of Generative AI Assistance}
A large language model (Claude Opus 4.8) was used for editorial assistance in condensing, anonymizing, and reformatting an earlier manuscript draft. The authors reviewed and edited the resulting text and remain responsible for the submission.

\end{document}